\documentclass[twocolumn,10pt]{asme2e}

\usepackage{graphicx}
\usepackage{mathtools}
\usepackage{amsfonts}
\usepackage{calrsfs}
\usepackage{bbm}
\usepackage{amsmath}
\usepackage{amssymb}
\usepackage{multirow}
\usepackage{makecell}
\DeclareMathAlphabet{\pazocal}{OMS}{zplm}{m}{n}

\usepackage{booktabs}
\usepackage{breqn}
\usepackage{xcolor}
\usepackage{hyperref}

\usepackage{float}

\usepackage{rotating}
\usepackage{multirow}
\usepackage{multicol}

\usepackage[utf8]{inputenc}

\usepackage{graphicx}
\usepackage{amsmath}

\usepackage[version=4]{mhchem}
\usepackage{siunitx}
\usepackage{longtable,tabularx}
\usepackage{bm}
\usepackage[inline]{enumitem}
\usepackage{array}

\confshortname{IDETC/CIE2021}
\conffullname{the ASME 2021 \\
International Design Engineering Technical Conferences \\
              and Computers and Information in Engineering Conferences}
\papernum{DRAFT DETC2021-xxxxx}
\confdate{17-20}
\confmonth{August}
\confyear{2021}
\confcity{Virtual Conference}
\confcountry{USA}

\papernum{DETC2021-xxxxx}

\title{RANGE-GAN: RANGE-CONSTRAINED GENERATIVE ADVERSARIAL NETWORK FOR CONDITIONED DESIGN SYNTHESIS}

\author{Amin Heyrani Nobari
    \affiliation{
	Dept. of Mechanical Engineering\\
	Massachusetts Institute of Technology\\
	Cambridge, MA 02139\\
    Email: ahnobari@mit.edu
    }	
}
\author{Wei Chen
    \affiliation{
	Siemens Technology\\
	Princeton, NJ 08540\\
    Email: chen.wei@siemens.com
    }	
}
\author{Faez Ahmed
    \affiliation{
	Dept. of Mechanical Engineering\\
	Massachusetts Institute of Technology\\
	Cambridge, MA 02139\\
    Email: faez@mit.edu
    }	
}

\newcommand{\eg}{{\em e.g.}}
\newcommand{\etal}{{\em et~al.}}
\newcommand{\ie}{{\em i.e.}}

\newcommand{\RNum}[1]{\uppercase\expandafter{\romannumeral #1\relax}}

\usepackage[linesnumbered,ruled,vlined]{algorithm2e} 

\begin{document}

\maketitle    

\begin{abstract}
{\it 
Typical engineering design tasks require the effort to modify designs iteratively until they meet certain constraints, \ie, performance or attribute requirements. Past work has proposed ways to solve the inverse design problem, where desired designs are directly generated from specified requirements, thus avoid the trial and error process. Among those approaches, the conditional deep generative model shows great potential since 1)~it works for complex high-dimensional designs and 2)~it can generate multiple alternative designs given any condition.
In this work, we propose a conditional deep generative model, Range-GAN, to achieve automatic design synthesis subject to range constraints. The proposed model addresses the sparse conditioning issue in data-driven inverse design problems by introducing a label-aware self-augmentation approach. We also propose a new uniformity loss to ensure generated designs evenly cover the given requirement range.
Through a real-world example of constrained 3D shape generation, we show that the label-aware self-augmentation leads to an average improvement of 14\% on the constraint satisfaction for generated 3D shapes, and the uniformity loss leads to a 125\% average increase on the uniformity of generated shapes' attributes.
This work laid the foundation for data-driven inverse design problems where we consider range constraints and there are sparse regions in the condition space.
}
\end{abstract}



\setlength{\belowdisplayskip}{5pt} \setlength{\belowdisplayshortskip}{5pt}
\setlength{\abovedisplayskip}{5pt} \setlength{\abovedisplayshortskip}{5pt}



\section{Introduction}
A typical design process involves tedious trial and error procedures, wherein a designer modifies design configurations based on functional, geometric, or aesthetics requirements. From the structural or materials design in engineering applications to the shape design of everyday objects, this design process usually takes considerable time and effort. To reduce human effort, one can automate the trial and error process by using optimization techniques, but the computational time scales up with the problem dimension~\cite{Bellman34}. Besides, instead of isolating a final optimal solution, one may be interested in discovering multiple alternatives. Recent advances in deep generative models allow people to train a data-driven model that can propose new design alternatives~\cite{burnap2016estimating,yang2018microstructural,oh2019deep,guo2019circuit,zhang20193d,chen2020airfoil,shu20203d,wang2020deep,chen2021padgan,chen2021deep}. A generative model learns the distribution of past exemplars so that one can draw new samples from that distribution. In most scenarios, rather than drawing random samples, we also want these new designs to have certain desired properties, such as particular attributes or performance requirements. While one can still use an optimization process to select desired candidates~\cite{yang2018microstructural,chen2020airfoil,chen2021deep}, this method has the same issues as mentioned earlier. Instead, a conditional generative model can be trained to learn a conditional distribution (\ie, distribution of designs conditioned on any property), so that new design candidates will be generated with given user-specified properties~\cite{liu2018generative,ma2019probabilistic,yilmaz2020conditional,achour2020development}. In this paper, we adopt this method and further address the problem of conditioning on range constraints~\textemdash~\ie, given upper and lower bounds on any property, how can we generate designs for which this property falls within the specified bounds. This is different from previous work, where conditions are set as exact values. The consideration of range constraints is more practical since, in many cases, some tolerance is allowed for certain properties. 
\begin{enumerate}
    \item How to enforce range constraints (with arbitrary lower/upper bounds) on generated samples?
    \item How to ensure accurate conditioning when data with some conditions are sparse or unavailable?
    \item How to allow the conditions of generated samples to uniformly fall within the range constraints?
\end{enumerate}

To address the above challenges, we propose a deep generative model called the \textit{range-constrained generative adversarial network (Range-GAN)}. We then demonstrate the effectiveness of our proposed model on a 3D shape design example. Our primary contributions are as follows:
\begin{enumerate}
    \item A new range loss function which allows for effective range conditioning in continuous spaces, with over 80\% constraint satisfaction.
    \item We introduce a novel loss term to encourage uniform coverage of the condition (label\footnote{In this paper, the word label refers to the actual value of a design's performance or attribute.}) space inside the acceptable constraint range, and show that this loss leads to a 125\% average increase on the uniformity of generated samples' labels.
    \item We propose label-aware self-augmentation, which allows for augmenting the data such that the sparsely populated regions of the condition (label) space are populated. We show that this augmentation improves the real-world performance of Range-GAN notably (by up to 18\%), which proves the effectiveness of the augmentation in improving Range-GAN.
    \item We apply Range-GAN in a 3D shape generation process to demonstrate the ease with which Range-GAN can be effectively applied to any GAN approach.
\end{enumerate}

\section{Background and Related Work}

Our work addresses an inverse design problem, where a designer inputs some requirements and expects an algorithm/model to generate desired designs. In this section, we first introduce the techniques used in the inverse design problem, especially the use of data-driven methods. Then, we provide a brief description of conditional generative adversarial networks, upon which our proposed model is based.

\subsection{Inverse Design Problem}

Inverse design problems have been studied in various engineering design domains such as airfoil shape design, mechanism design, and metamaterial design. A typical inverse design problem can be solved by optimization, where we optimize design parameters such that the design's performance satisfies certain objectives or constraints. Unfortunately, gradient-based optimization (\eg, topology optimization or adjoint-based shape optimization) is restricted to limited design representations and solver types. On the other hand, in gradient-free optimization (\eg, genetic algorithm or Bayesian optimization), as the problem dimension (\ie, the number of design parameters) increases, the computational cost quickly becomes prohibitive due to the curse of dimensionality~\cite{Bellman34}. Alternatively, one can use a neural network as a surrogate for any physics simulation, so that standard back propagation can be used to get analytical gradients for gradient-based optimization~\cite{peurifoy2018nanophotonic}. This approach applies to any black-box physics solvers and does not have the computational cost problem seen in gradient-free optimization methods.

While many traditional techniques on the inverse design problem focus on the use of optimization, research has been done to completely remove such time-consuming iterative optimization processes to significantly reduce the computational cost. 
Reinforcement learning (RL) offers one way to achieve this goal.
For example, Vermeer~\cite{vermeer2018kinematic} used temporal difference (TD) learning to synthesize mechanism designs with desired output trajectories. Lee~\etal~\cite{lee2019case} used a DQN to design a microfluidic device which led to a target flow shape.
While RL works well for discrete and low-dimensional design spaces, it does not effectively scale to more complex scenarios~\cite{theodorou2010reinforcement}. 

The aforementioned approaches only allow for bijective mappings between the target and design parameters, which might be impractical for many problems. For example, a specific lift or drag coefficient may correspond to multiple feasible design solutions. This becomes more obvious when we only require that the target performance falls within a range rather than at an exact point. Recent advances in deep generative models, generative adversarial networks (GANs)~\cite{goodfellow2014generative}, and variational autoencoders (VAEs)~\cite{kingma2013auto} in particular, provide ways to solve this problem. Deep generative models can learn a distribution of designs so that one can quickly sample plausible designs without any optimization process. The learned distribution can be further conditioned on target performance by using models like conditional GANs (cGANs)~\cite{mirza2014conditional} or conditional VAEs (CVAEs)~\cite{sohn2015learning}. This allows us to generate many designs conditioned on any performance requirement; \ie, the mapping from the performance space to the design space can be one-to-many. This technique has been used for the inverse design of metasurfaces, metamaterials, and cellular structures~\cite{liu2018generative,ma2019probabilistic,wang2021ih}. 
As pointed out in~\cite{ding2020ccgan}, however, conditional generative models with continuous conditions may fail. Unlike finite discrete conditions, design data under certain continuous performance conditions can be sparse or even non-existing, leading to inaccurate conditioning under those conditions and subsequent failure of the model. Past work~\cite{yilmaz2020conditional,achour2020development} proposes to discretize the continuous values of the metrics. This approach, however, eliminates the possibility of setting arbitrary conditions.

In this paper, we address the above issue in continuous conditional generative models. Moreover, we consider a more general problem, where, instead of using exact performance metrics as conditions, we generate designs with performances within any user-defined range. 
We achieve this goal by modifying the cGAN model, which we introduce in the next section.

\subsection{Conditional Generative Adversarial Networks}

A generative adversarial network~\cite{goodfellow2014generative} consists of two models~\textemdash~a \textit{generator} and a \textit{discriminator}. The generator $G$ maps an arbitrary noise distribution to the data distribution, in our case the distribution of designs, and can thus generate new data; simultaneously, the discriminator $D$ tries to perform classification (\ie, attempts to distinguish between real and generated data). Both models are usually built with deep neural networks. As $D$ improves its classification ability, $G$ also improves its ability to generate data that fools $D$. 
Thus, a vanilla GAN (standard GAN with no bells and whistles) has the following objective function:
\begin{equation}
\min_G\max_D V(D,G) = \mathbb{E}_{\mathbf{x}\sim P_{data}}[\log D(\mathbf{x})] + 
\mathbb{E}_{\mathbf{z}\sim P_{\mathbf{z}}}[\log(1-D(G(\mathbf{z})))],
\label{eq:gan_loss}
\end{equation}
where $\mathbf{x}$ is sampled from the data distribution $P_{data}$, $\mathbf{z}$ is sampled from the noise distribution $P_{\mathbf{z}}$, and $G(\mathbf{z})$ is the generator distribution. A trained generator can thus map from a predefined noise distribution to the distribution of designs.

The \textit{conditional GAN}, or \textit{cGAN}~\cite{mirza2014conditional}, further extends GANs to allow the generator to learn a conditional distribution. This is done by feeding the condition, $\mathbf{y}$, to both $D$ and $G$. The loss function then becomes:
\begin{equation}
\begin{split}
\min_G\max_D V_{\text{cGAN}}(D,G) = & \mathbb{E}_{\mathbf{x}\sim P_{data}}[\log D(\mathbf{x}|\mathbf{y})] + \\
& \mathbb{E}_{\mathbf{z}\sim P_{\mathbf{z}}}[\log(1-D(G(\mathbf{z}|\mathbf{y})))].
\end{split}
\label{eq:cgan_loss}
\end{equation}

Therefore, given any conditions, cGAN can generate a set of designs that satisfy the given conditions, by feeding a set of random noise. In this paper, we use range constraints as conditions~\textemdash~\ie, the performance/attribute of generated designs needs to fall within some lower and upper bounds. This is a more practical consideration because a certain level of tolerance on the performance/attribute is allowed in many cases.

As mentioned in the previous section, conditional GAN may fail when the conditions are continuous due to data sparsity at certain conditions. To address this issue, past work either discretizes continuous conditions~\cite{yilmaz2020conditional,achour2020development} or proposes a new sampling scheme to mitigate the problem~\cite{ding2020ccgan}. However, neither method works well with large sparse regions in the condition space. In this paper, we address this problem by using a label-aware self-augmentation method, which we elaborate on in Sect.~\ref{sec:augment}.

\subsection{3D Shape Synthesis via Deep Generative Models}

In this paper, we validate our proposed model on a 3D shape synthesis task. The goal of this task, in general, is to generate useful new 3D shapes while avoiding manual efforts/expertise to construct and modify detailed geometries. 
Deep generative models, like GANs or VAEs, are excellent candidates for this task due to their ability to learn complex data distributions and generate realistic samples. The model architecture, cost, and performance are highly dependent on data representation. Volumetric representations, like voxel grids and point clouds, are straight-forward to learn but require large models and hence have wasteful computational or memory cost~\cite{wu2016learning,wang2018global,achlioptas2018learning,liu2018learning,zhang20193d,shu20203d}. View-based approaches generate multi-view depth maps, normal maps, or silhouettes, which reduce the computational/memory cost, but cannot produce shapes with self-occlusion~\cite{arsalan2017synthesizing}. Surface patch representation uses one or more images to represent the 3D object’s surface. This allows self-occlusion but requires complex data preprocessing and shape correspondence~\cite{ben2018multi}. To further reduce computation/memory requirement, 3D objects are represented as implicit fields (\eg, signed distance fields) and neural networks are trained to approximate those functions, simply mapping 3D coordinates to scalars~\cite{mescheder2019occupancy,chen2019learning,park2019deepsdf}. This produces simple neural network architectures and generates shapes with no resolution limit, since each shape is represented as a continuous implicit field. 

Our model for 3D shape synthesis is built on IM-NET~\cite{chen2019learning}, which is one of the implicit field-based methods. The IM-NET consists of two parts: representation learning (\ie, learning a latent vector representation for 3D shapes) and generative modeling (\ie, learning the latent vector distribution and generating new latent vectors). We modify the second part such that the latent vector distribution can be conditioned on any range constraints. We will further elaborate our shape generation pipeline in Sect.~\ref{sec:pipeline}.

\section{Methodology}
\label{sec:methodology}
\begin{figure*}[ht!]
\centering
\vskip -0.4in
\includegraphics[width=1.7\columnwidth]{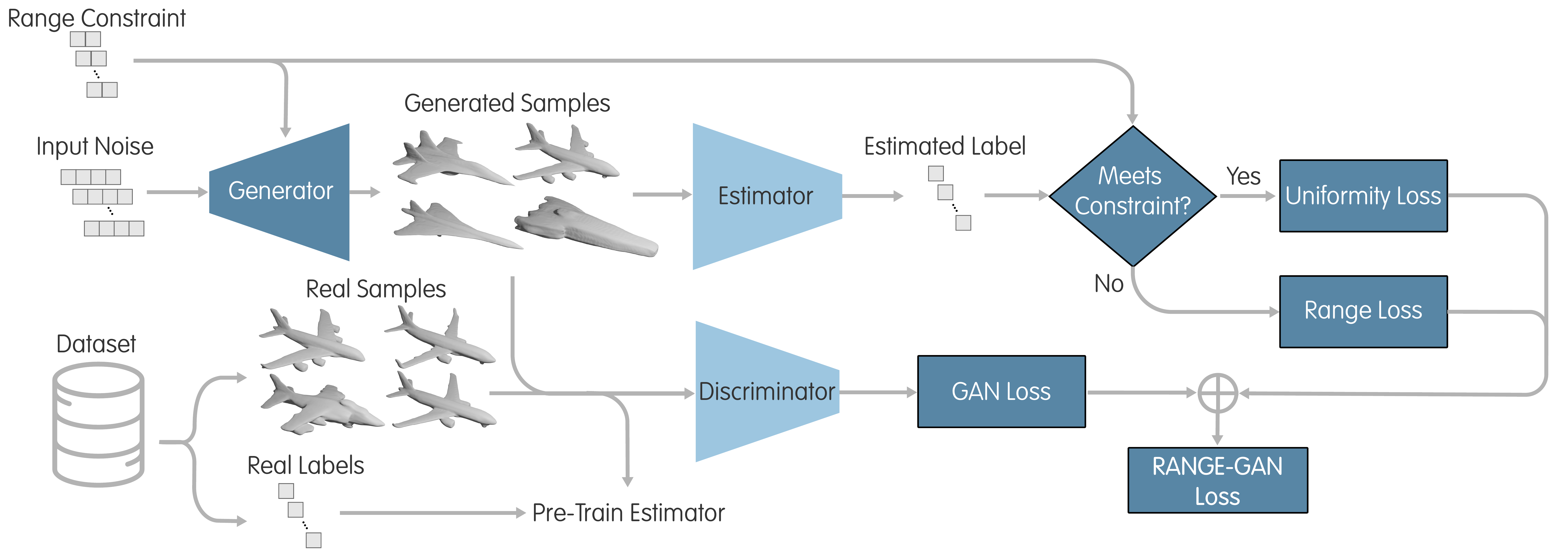}
\vskip -0.1in
\caption{Overall architecture for the Range-GAN}
\label{fig:architecture}
\end{figure*}

This section describes our overall methodology. We start by discussing the overall pipeline for 3D shape generation (Sect.~\ref{sec:pipeline}). Next, we propose new loss functions and generator architecture to effectively enforce range-constraints (Sect.~\ref{sec:range}). The following two sections address problems with range-constraint conditioning: Sect.~\ref{sec:unform} addresses the problem of getting uniformly distributed samples within a range, while Sect.~\ref{sec:augment} addresses data sparsity with augmentation methods. Our methodology concludes by showing how these methods apply to constraints on multiple variables. We summarize the overall architecture of Range-GAN in Fig.~\ref{fig:architecture}. 

\subsection{3D Shape Generation Pipeline}
\label{sec:pipeline}
While our method can be applied to various use cases, we demonstrate the effectiveness of the Range-GAN on a 3D shape design example. To generate 3D models, we combine our Range-GAN with the \textit{implicit decoder} in the \textit{IM-NET}~\cite{chen2019learning}. Specifically, the implicit decoder is a neural network that approximates the implicit fields of shapes. At each point $(x, y, z)$, an implicit field assigns a value of 1 if the point is inside the 3D shape, and 0 otherwise. Given the implicit field of a shape, we can obtain its surface mesh by using methods like marching cubes~\cite{lorensen1987marching}. Each implicit field is also conditioned on a \textit{latent vector}, which represents a unique shape and is learned by an encoder. The implicit decoder takes in both a latent vector and a 3D point coordinate and predicts the implicit field at the given point to describe a shape represented by the given latent vector. The combination of the encoder and the implicit decoder is called an \textit{IM-AE} and is trained according to Ref.~\cite{chen2019learning}. After training, we obtain the latent vectors for the training data. We can further use our Range-GAN to learn the distribution of those latent vectors. By concatenating the trained generator of the Range-GAN with the trained implicit decoder, we can generate 3D shapes based on any condition and noise vector, as shown in Fig.~\ref{fig:workflow} (we will elaborate on the generator training in the following sections). The merit of using this framework is that we can modularize two tasks~\textemdash~1)~generative modeling of latent vectors and 2)~transformation from latent vectors to 3D shapes. This, on one hand, simplifies the generative modeling task, because the GAN is learning a much lower-dimensional distribution compared to directly learning the distribution of 3D shapes. On the other hand, this provides a platform for future study on generative modeling for 3D shapes, where we do not need to worry about shape representation, because we already have the latent vector as an efficient shape representation.

\subsection{Enforcing Range Constraints in GANs}
\label{sec:range}

Similar to the idea of adding an auxiliary classifier/regressor to the cGAN~\cite{odena2017conditional,wang2021ih} to improve generation quality or enforce accurate conditioning, we use a label estimator to guide the generator during training to generate designs that meet input constraints. In this way, we utilize the discriminator's insight into the data distribution to promote synthesis of realistic designs while simultaneously using the estimator's insight to guide the generator towards meeting input requirements. In this approach, the estimator can be any differentiable model that predicts the label of generated designs (\eg, an adjoint-based physics simulator or a deep neural network~(DNN) regression model). In this paper, we use a pre-trained DNN-based estimator to predict the labels of any given design. To integrate the estimator into the GAN's objective, we propose a novel loss function for the generator. This loss function must have certain characteristics to be effective~\textemdash~1)~the loss function must have a zero gradient for samples within the input condition range~(\ie, samples that meet the condition need no further change) and 2)~the gradient should start gradually decreasing as samples get closer to the acceptable range to stabilize training. Given that these characteristics are seen in the negative log likelihood~(NLL) function of the GAN objective, we attempt to create a similar loss function that applies the same principles for the range conditions. To imitate the NLL, we need a mechanism that turns predicted continuous labels into probabilities of condition satisfaction. To do this, we use two sigmoid functions shifted to the lower and upper bounds of the given range condition to estimate the probability of condition satisfaction:
\begin{equation}
\begin{split}
p(\mathbf{x}|[\mathbf{y}_{lb},\mathbf{y}_{ub}])	\approx\frac{1}{1+e^{\phi(E(\mathbf{x})-\mathbf{y}_{lb})}}-\frac{1}{1+e^{\phi(E(\mathbf{x})-\mathbf{y}_{ub})}},
\end{split}
\label{eq:3}
\end{equation}
where $E(\mathbf{x})$ is the label predicted by the estimator $E$, $\mathbf{y}_{lb}$ and $\mathbf{y}_{ub}$ are the lower and upper bounds of the range constraint, and $\phi$ is a scaling factor determining how aggressively the probability grows/shrinks at the lower/upper bounds. The hyper-parameter $\phi$ determines how aggressive the overall gradient will be, and how close the sigmoids will be to a unit step function. We measure the NLL based on this estimated probability using the \textit{range loss function}:
\begin{equation}
\begin{split}
\mathcal{L}_{\mathrm{range}}=-\frac{\sum_{i=1}^{N}\mathbbm{1}_{{(y_{i}-y_{i,ub})	\times(y_{i}-y_{i,lb})\geq 0}} \,\log(p(\mathbf{x}_{i}|[y_{i,lb},y_{i,ub}]))}{\sum_{i=1}^{N}\mathbbm{1}_{{(y_{i}-y_{i,ub})	\times(y_{i}-y_{i,lb})\geq 0}}}.
\end{split}
\label{eq:4}
\end{equation}
Here, $N$ is the number of samples generated in a batch and $\mathbf{x}_{i}$ is the $i$-th sample in the batch with an estimated/calculated label of $\mathbf{y}_{i}$. Note that $\mathcal{L}_{\mathrm{range}}=0$ for samples that meet the condition ($y_{i,lb}\leq y_{i}\leq y_{i,ub}$). When this loss term is added to the generator loss, the estimator will guide the generator to produce samples that satisfying the range condition as training proceeds.

We do not have $\mathbf{y}_{lb}$ and $\mathbf{y}_{ub}$ in the data. During training, $y_{i,lb}$ and $y_{i,ub}$ are randomly sampled as follows:
\begin{equation}
\begin{split}
&y_{i,lb}\sim \text{unif}(y_{\min}, y_{\max}-0.05(y_{\max}-y_{\min})),\\
&y_{i,ub}\sim \text{unif}(y_{i,lb}+0.05(y_{\max}-y_{\min}), y_{\max}),
\end{split}
\end{equation}
where $y_{\min}=\min_i \{y_i\}$ and $y_{\max}=\max_i \{y_i\}$. When we scale the label to the range [0,1], we simply have $y_{\min}=0$ and $y_{\max}=1$. 
We only sample one condition at every training step and use that condition for the entirety of the batch. We found this treatment gave better results. Also, as we will discuss in Sect.~\ref{sec:unform}, this is a necessary requirement for our uniformity loss.

\subsection{Incorporating Conditions by Conditional Batch Normalization}

\begin{figure*}[ht!]
\centering
\vskip -0.4in
\includegraphics[width=1.8\columnwidth]{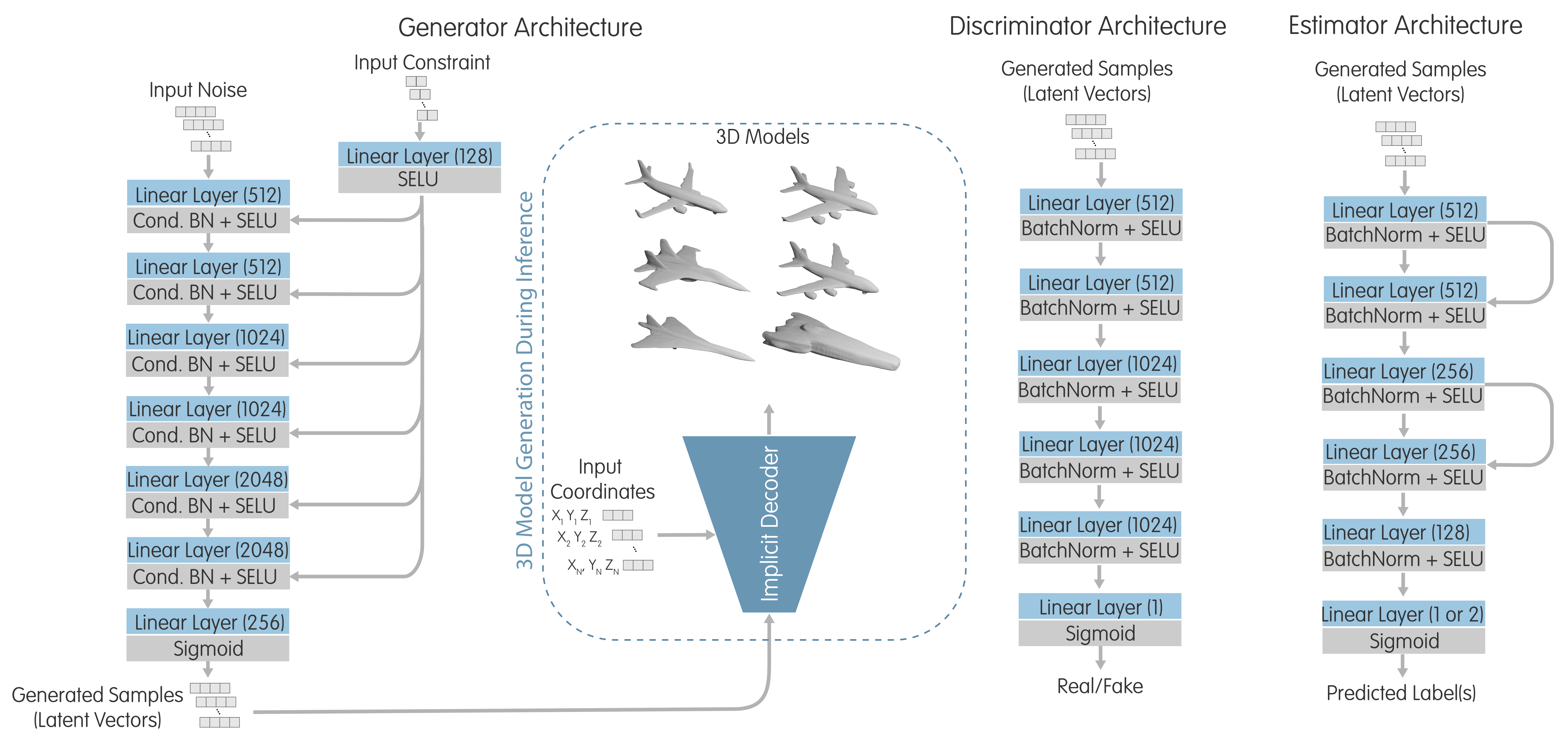}
\vskip -0.1in
\caption{3D shape generation pipeline during inference (\textit{left}) and detailed architectures for the generator (\textit{left}), the discriminator (\textit{middle}), and the estimator (\textit{right}). The residual connections in the estimator take the outputs of the higher layer before activation and batch normalization and add them to the outputs of the following layer before activation and batch normalization. (Cond. BN: conditional batch normalization; SELU: scaled exponential linear units~\cite{selu})}
\label{fig:workflow}
\end{figure*}

A na\"ive approach to incorporate input conditions in cGANs is to concatenate input conditions with the input noise vector before feeding them into the generator~\cite{mirza2014conditional}. For categorical labels, which cGANs were developed for, this approach is effective as input labels are typically one-hot embeddings that are highly discrete and distinct. On the other hand, scalar labels in continuous spaces may not be very distinct and are therefore not suitable for this approach~\cite{ding2020ccgan}. We avoid this issue by applying conditional batch normalization~\cite{cbn} to the output of every linear layer in the generator (Fig.~\ref{fig:workflow}), where the conditional batch normalization is computed based on the input conditions. This approach is effective in feeding continuous conditions~\cite{ding2020ccgan}.

\subsection{Enforcing Label Uniformity} 
\label{sec:unform}

It is important to note that even the labels of generated designs satisfy a given range constraint, the label distribution can vary. For example, labels can either spread uniformly over the entire range or gathered at one point. So far in our implementation of the range loss~(Eqn.~\ref{eq:4}), samples that meet any given condition are treated equally, with no gradient applied to them based on their predicted labels. In this paper, we introduce an approach to promote uniform coverage of the condition range. Uniform coverage is the most generalizable case, because, since the generator can be trained to cover all of the condition space, it should theoretically also be possible to bias the generator towards the lower or higher bounds of any given input condition. Uniform coverage demonstrates that the generator can cover an entire range space and biasing the generator towards either bound should be as simple as changing the loss function to encourage the generator towards either bound.

To promote uniformity in the labels of generated designs, we can maximize any entropy-based metric on those labels. In our experiments, we found that general entropy losses are not very useful, as they encourage overall entropy, which means the loss terms encourage the generator to generate samples with labels having the largest possible distance, thus pushing generated designs outside the constraint range and significantly decreasing condition satisfaction rate. To overcome this, we introduce a novel \emph{uniformity loss} which takes into account the acceptable range and encourages entropy only within that acceptable range. Furthermore, our proposed uniformity loss is particularly geared around encouraging uniform distribution and takes advantage of a specific property of uniform distributions. 

Specifically, we take advantage of two properties: 1)~given a starting distribution that is uniform, if we split this distribution into two, both resulted distributions will also be uniform; 2)~the mean of any uniform distribution is the mean of the lower and upper bound of the distribution. With those properties, the uniformity loss slices the generated samples' label distribution at random points within the constraint range. It then measures the mean of each label subset and uses the mean absolute error between the actual mean and the expected uniform distribution mean as the loss term. This is done multiple times for each batch during training to ensure that uniformity is stochastically encouraged. This is because, theoretically, if the slice points remained constant, a non uniform multi-modal distribution could mimic a uniform distribution for a set of slice points, where the modes of said distribution are located exactly between slice points. This loss function will only be applied to the samples that meet the input condition and can be formulated as such:
\begin{equation}
\begin{split}
&\mathcal{L}_{\mathrm{unif}}=\frac{1}{K}\\
&\sum_{j=1}^{K} \mathbb{E}_{\epsilon_{j} \sim \mathbb{U}(y_{lb}, y_{ub})}|\frac{
\sum_{i=1}^{N}\mathbbm{1}_{{(y_{i}-y_{ub})\times(y_{i}-\epsilon_{j})\leq 0}} \,(y_{i}-\frac{y_{ub}+\epsilon_{j}}{2})}{\sum_{i=1}^{N}\mathbbm{1}_{{(y_{i}-y_{ub})\times(y_{i}-\epsilon_{j})\leq 0}}}|+\\
&|\frac{
\sum_{i=1}^{N}\mathbbm{1}_{{(y_{i}-y_{lb})\times(y_{i}-\epsilon_{j})\leq 0}} \,(y_{i}-\frac{y_{lb}+\epsilon_{j}}{2})}{\sum_{i=1}^{N}\mathbbm{1}_{{(y_{i}-y_{lb})\times(y_{i}-\epsilon_{j})\leq 0}}}|,
\end{split}
\label{eq:5}
\end{equation}
where we apply random splits $K$ times using slice points~($\epsilon_{j}$) uniformly sampled in the input condition range. Note that we use the entire batch to form the distribution. Therefore, $y_{ub}$ and $y_{lb}$ are sampled once per batch and are used to generate every sample in that batch during training. At this point, the overall objective of Range-GAN can be written as the combination of the vanilla GAN objective from Eqn.~\ref{eq:gan_loss} and the new loss terms:
\begin{equation}
\min _{G} \max _{D} V(D, G)+\lambda_{1} \mathcal{L}_{\mathrm{range}}(G) +\lambda_{2} \mathcal{L}_{\mathrm{unif}}(G),
\end{equation}
where hyper-parameters $\lambda_{1}$ and $\lambda_{2}$ determine the weight of the range and uniformity losses, respectively.

\subsection{Label-Aware Self-Augmentation to Address The Data Sparsity Problem} \label{sec:augment}

As discussed before, sometimes exact estimators of any design's label~(\eg, performance metrics, attributes, etc.) are not readily differentiable. In these circumstances, the estimator can be a pre-trained DNN regression model which predicts the label based on a set of training data. It is, however, often the case that the labels in the data do not evenly cover the label space. In these circumstances, there may not be enough data associated with certain regions of the label space for the DNN estimator to learn those regions with high accuracy. This leads to inconsistencies in the actual labels of the generated designs compared to the predicted labels in the sparsely populated regions of the label space. This issue is often seen in engineering design datasets, where the extremes of any given label do not have many samples associated with them. 

In this paper, we propose a self-augmentation method that uses the generated samples to augment the dataset and retrain the DNN estimator to better cover sparse regions of the label space. Specifically, after the first round of training is finished on the GAN, a number of samples are generated and evaluated using a high fidelity \textit{evaluator} (\eg, a physics simulator). Unlike the label estimator mentioned earlier, such an evaluator is not necessarily differentiable and can be any black-box model. We then add samples from this subset to the dataset if their actual labels are located in sparse regions. Particularly, we split the label space into 10 equally spaced bins and count the number of samples from the data in each bin. We then add data from the generated and evaluated samples to bins with a smaller number of samples until the bin counts become equal or no more samples exist in the newly generated set to fill the less populated bins. We then re-train the DNN estimator using this augmented dataset and use the improved DNN estimator and the newly generated and evaluated data to re-train the GAN. By doing this, we overcome the initial shortcomings of the DNN in sparsely populated regions of the label space, which will, in turn, guide the generator towards meeting input conditions better. Generally, the quality of the estimator is crucial to Range-GAN's success, and exact differentiable estimators will ultimately be the strongest option. Regardless, the lack of data in certain regions will also always impact the performance of the trained GAN, given the fact that, if data doesn't exist in certain regions, GANs cannot be expected to explore those regions. Ultimately, GANs only emulate the dataset and do not typically generate novel samples.



\section{Experimental Settings}

In this paper, we illustrate our results using the real-world example of synthesizing 3D airplanes. For this purpose, we use the airplane subset of the ShapeNet dataset~\cite{chang2015shapenet}, which includes 4,043 airplane models. We then measure the \textit{aspect ratio} and \textit{volume ratio} of models and use them as the labels for conditioning. The aspect ratio is the measure of the ratio between fuselage length and wingspan, while the volume ratio is the ratio of the volume of the 3D model compared to a unit cube it occupied. And since the volume of the unit cube is constant, the volume ratio practically indicates the volume of the model.

\begin{figure}[ht!]
\centering
\vskip -0.1in
\includegraphics[width=0.8\columnwidth]{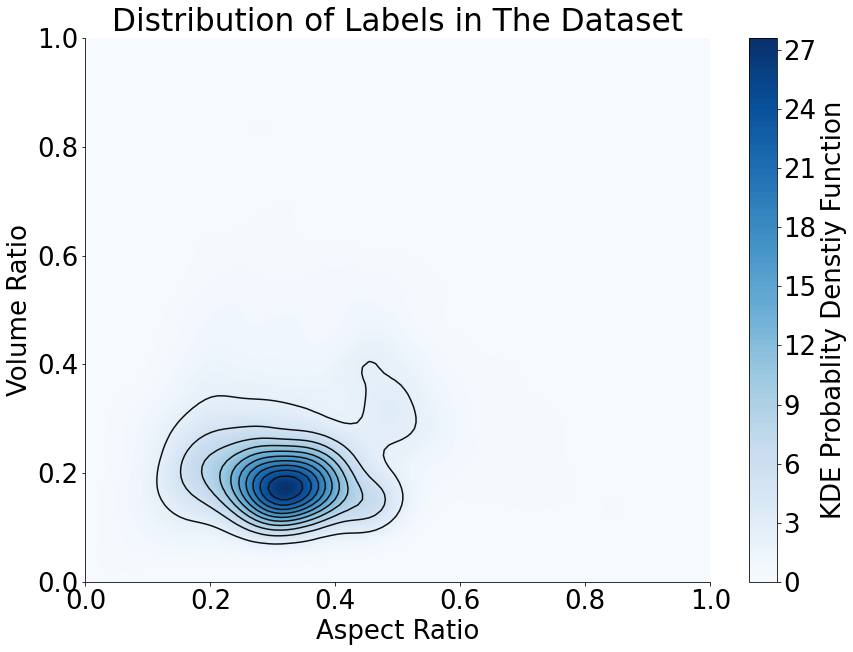}
\vskip -0.1in
\caption{A kernel density estimation of the probability density function for the labels in the dataset, demonstrating the distribution of labels within the dataset. Note that these labels report the normalized values of volume and aspect ratios, and not the true values.}
\vskip -0.1in
\label{fig:overall_dist}
\end{figure}

As discussed prior, sparsity in the label space is a major problem, which also exists in this dataset. To avoid extremely sparse sectors of the label space, we remove the samples beyond the 99.5 percentile in both volume and aspect ratios. After this step, we are left with a dataset of 4,012 models. The distribution of labels in this dataset is presented in Fig.~\ref{fig:overall_dist}. As evident, the data is mostly centered around a very narrow region of the label space, which makes this dataset a perfect example to demonstrate the effectiveness of Range-GAN even in such extremely sparse datasets.

\begin{figure}[ht!]
\centering
\includegraphics[width=1\columnwidth]{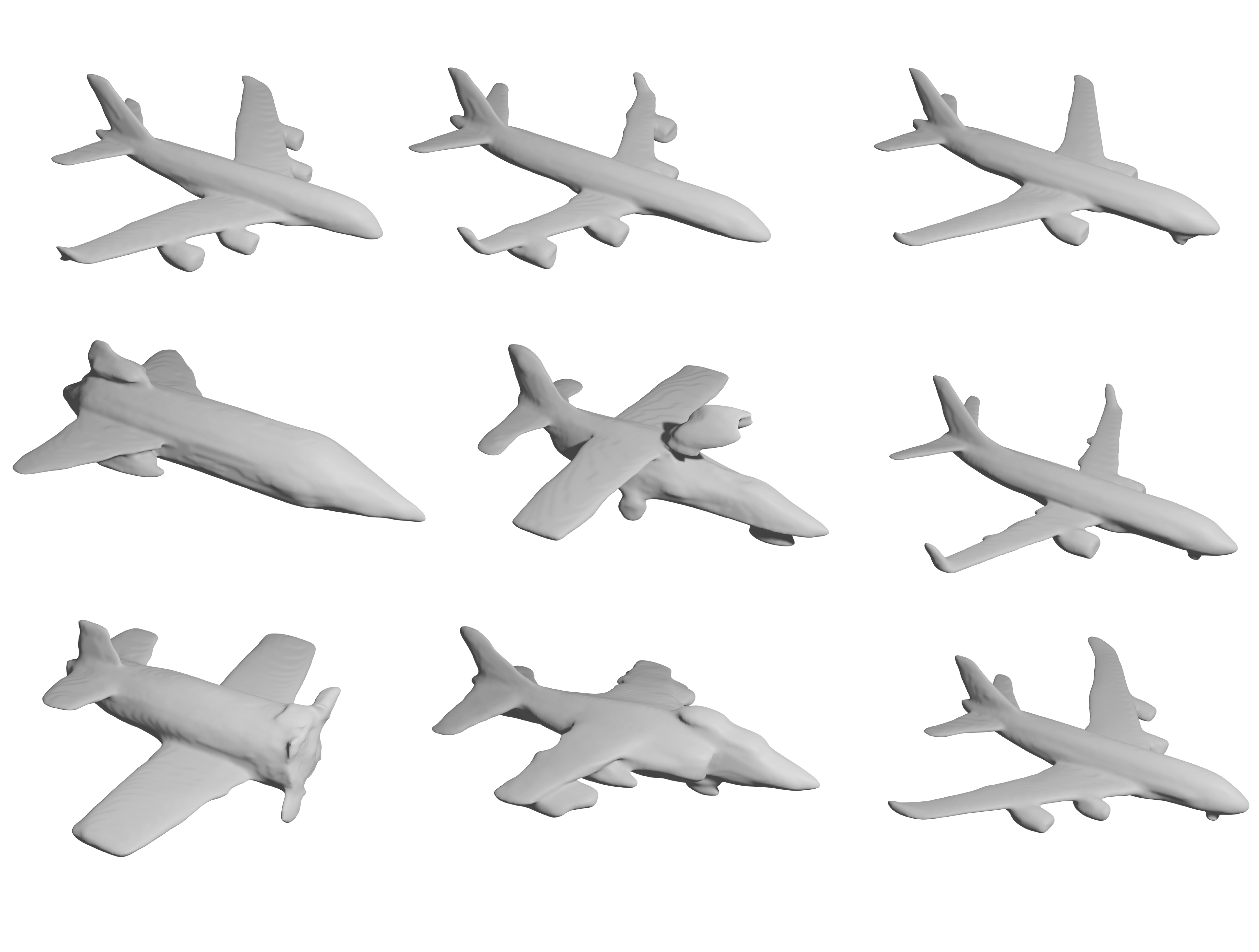}
\vskip -0.1in
\caption{A subset of the ShapeNet airplane dataset reconstructed by our trained IM-NET model.}
\vskip -0.1in
\label{fig:dataset}
\end{figure}

We then train the IM-NET model (as described in Sect.~\ref{sec:pipeline} and Ref.~\cite{chen2019learning}) on this dataset and use the 256-dimensional latent vectors to train Range-GAN. After training, the Range-GAN can generate new latent vectors, which can then be transformed into the implicit field representation of 3D models using the implicit decoder. We further obtain the surface meshes via marching cubes~\cite{lorensen1987marching}. Fig.~\ref{fig:dataset} demonstrates some examples of the ShapeNet airplanes reconstructed using the implicit decoder.

For all experiments, we set $\phi=20$, $\lambda_{1}=2.0$, and $\lambda_{2}=1.0$. 
We train our models for 50,000 steps with a batch size of 32. We use the Adam optimizer with a learning rate of $10^{-4}$, which decays by 20\% every 5,000 steps. We train the estimator using the L2 regression loss for 10,000 steps with a batch size of 256 using the Adam optimizer with a learning rate of $10^{-4}$, which decays by 40\% every 2,500 steps. To reduce bias in the discriminator, we sample the data during training such that the labels of any given batch are uniformly distributed across the label space. To do this, we uniformly sample random numbers between 0.0 and 1.0, which are the normalized bounds of the labels, and pick the sample with the label closest to the random number. We do the same when training the estimator; we found that this training improves the estimator significantly. 

\section{Results and Discussion}

In this section, we present the results of training Range-GAN based on the methods described above. First, we demonstrate Range-GAN in the context of single constraint conditioning for both the volume and aspect ratio labels independently. Then, we demonstrate the performance of Range-GAN on conditioning in both volume and aspect ratio labels simultaneously. In the end, we discuss the effects of our data augmentation and the performance of Range-GAN when measured based on exactly calculated labels. In sections before the augmentation, unless otherwise specified, the results presented are based on estimator predictions rather than exact values, which we do because calculating the actual labels is computationally expensive. Finally, we normalize the labels for both volume and aspect ratio to span the full range from 0.0 to 1.0. The values of aspect ratio (A/R) and volume ratio (V/R) presented here are based on the normalized label values and not the physical values. 

\subsection{Evaluation Metrics}
In this paper, the primary objective is for the generator to meet the input range condition. As such, we measure the success of any model by how well it can satisfy the input conditions. We do this using the condition satisfaction metric, which essentially measures the number of generated samples that meet the input condition:
\begin{equation}
\begin{split}
Satisfaction=\frac{\sum_{i=1}^{N}\mathbbm{1}_{{(y_{i}-y_{i,ub})	\times(y_{i}-y_{i,lb})\leq 0}}}{N},
\end{split}
\label{eq:4}
\end{equation}
where $N$ is the total number of generated samples.
The second metric we use is the measure of uniformity in the output distribution for the samples that meet the input condition. To do this, we use the quadratic entropy. Quadratic entropy is the mean of the square of the distances between any two samples' labels:
\begin{equation}
\begin{split}
Quadratic Entropy=\sum_{i=1}^{N}\sum_{i=j}^{N}\frac{1}{N^2}(y_{i}-y_{j})^2,
\end{split}
\label{eq:4}
\end{equation}
where $N$ is the number of generated samples that meet the input condition, and $y_{i}$ and $y_{j}$ are the labels of any two generated samples that meet the input condition.

\subsection{Single Constraint Case Studies}
\begin{figure*}[ht!]
\centering
\includegraphics[width=1\columnwidth,height=2.0in]{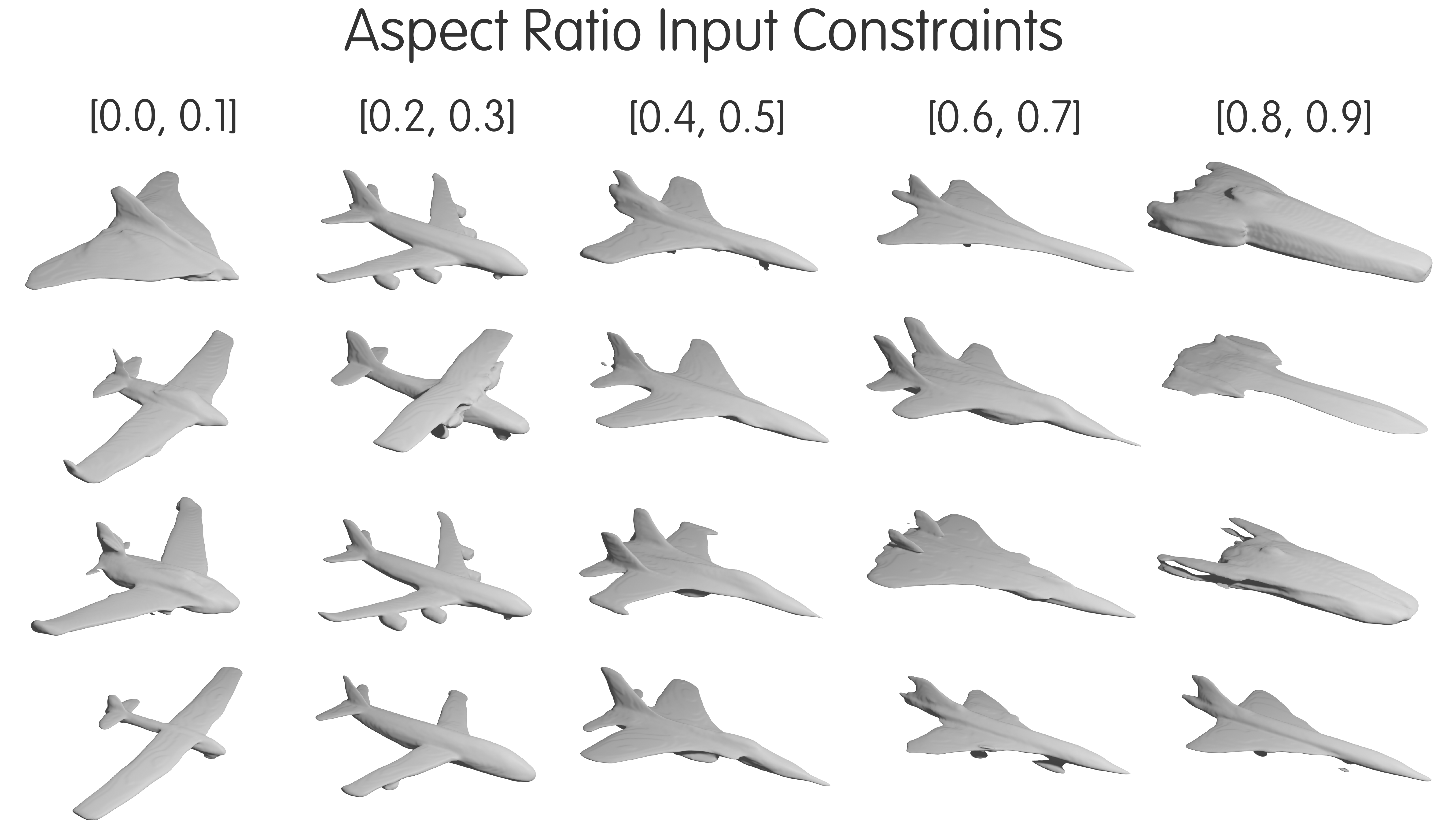}
\includegraphics[width=1\columnwidth,height=2.0in]{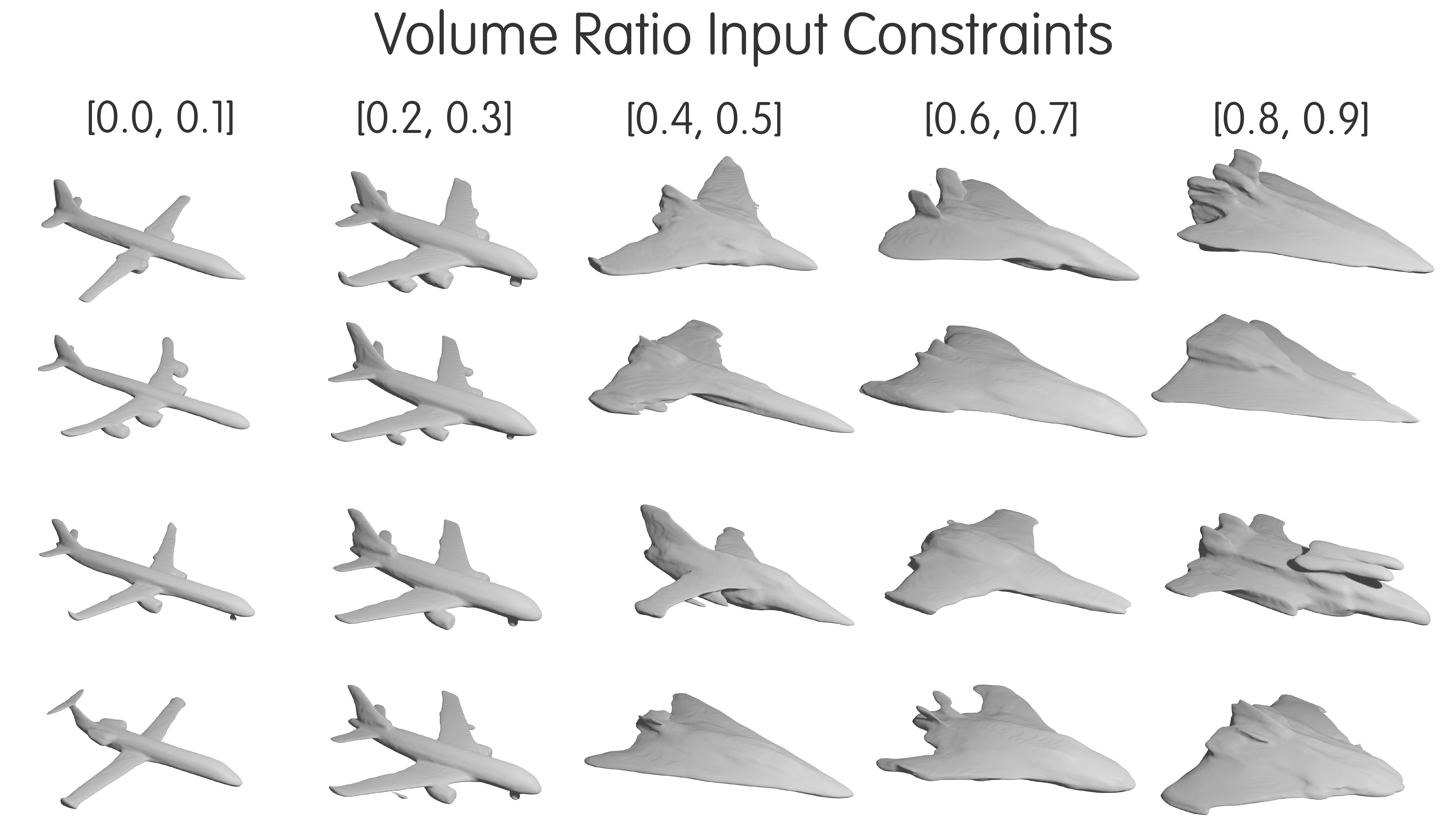}
\vskip -0.1in
\caption{Left: Example of Range-GAN outputs with different aspect ratio conditions. The input range conditions are presented above each column of images. The images in each column represent the outputs of Range-GAN given a single input condition. Right: Example of Range-GAN outputs with different volume ratio conditions. The input range conditions are presented above each column of images. The images in each column represent the outputs of Range-GAN given a single input condition. The images are to scale so models that appear larger occupy a larger volume.}
\label{fig:ratio_res}
\end{figure*}

\begin{figure}[ht!]
\centering
\includegraphics[width=1\columnwidth,height=2.0in]{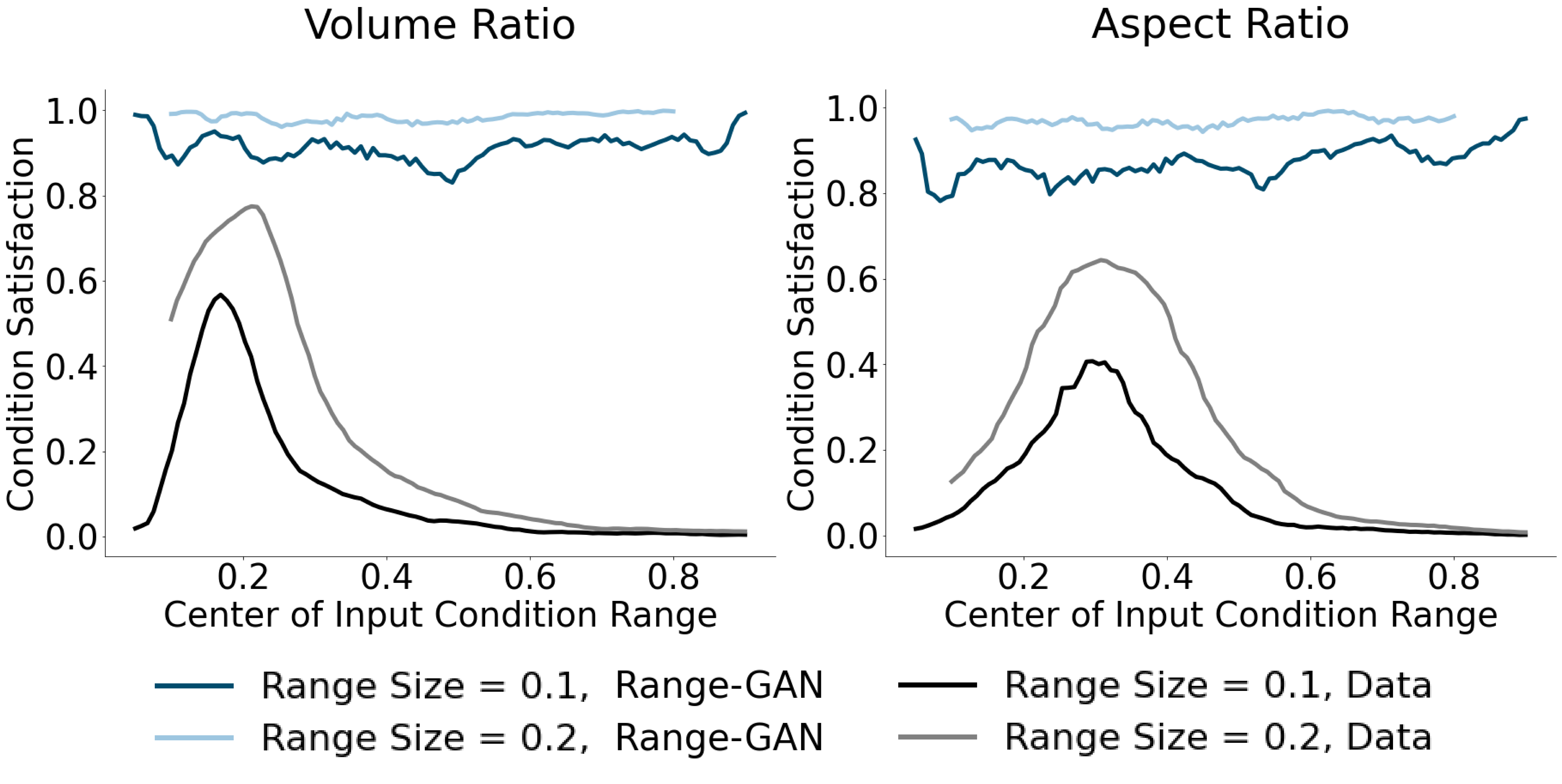}
\vskip -0.1in
\caption{The performance of Range-GAN in input condition satisfaction for both volume and aspect ratios compared to the same metric in the data distribution. The x-axis on the plots represents the center of the input range condition, meaning input condition of the range with width of $range size$ centered at the value on the x-axis.}
\vskip -0.1in
\label{fig:satisfaction}
\end{figure}

\begin{figure}[ht!]
\centering
\includegraphics[width=1\columnwidth]{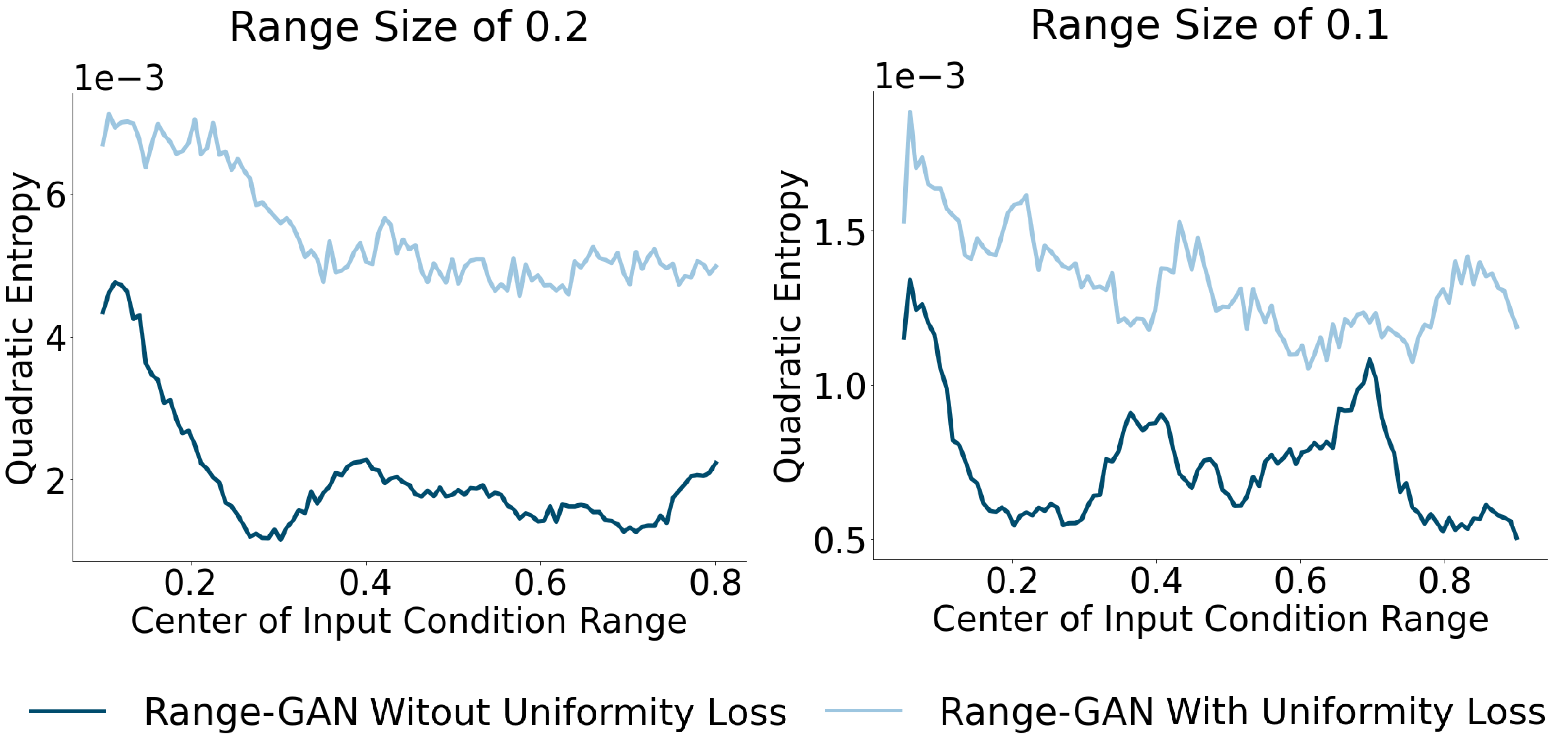}
\vskip -0.1in
\caption{The effect of the uniformity loss term on the output labels' quadratic entropy for two different range sizes at the input condition.The x-axis on the plots represents the center of the input range condition.}
\vskip -0.1in
\label{fig:uniformity}
\end{figure}

\begin{figure}[ht!]
\centering
\includegraphics[width=1\columnwidth]{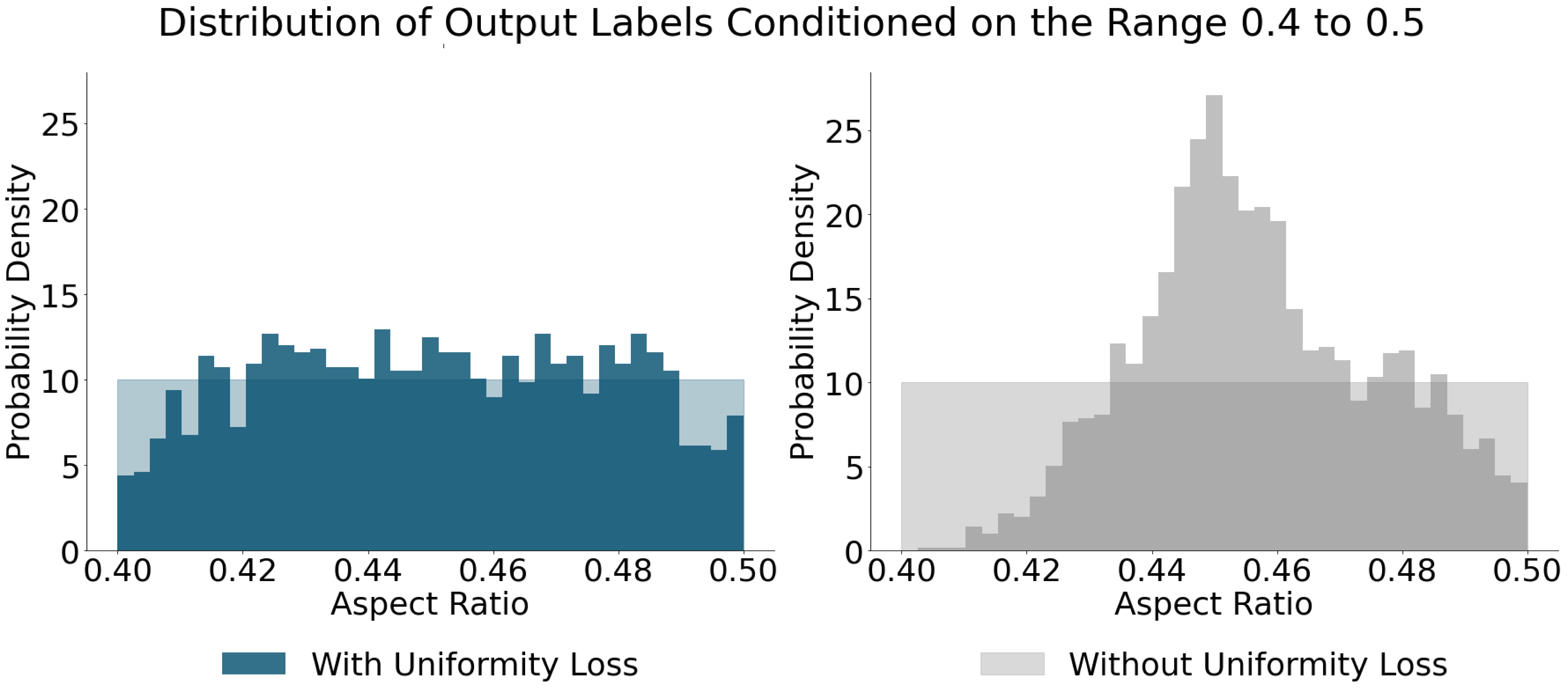}
\vskip -0.1in
\caption{The histograms of the distribution of output labels (for the aspect ratio case) in the acceptable range of 0.4 to 0.5~(these conditions were picked arbitrarily to visually demonstrate the distributions), both with and without the uniformity loss. Both figures represent probability density on the y-axis. The overlaid constant distribution at a probability density of 10 illustrates a perfectly uniform distribution.}
\label{fig:dist}
\end{figure}

First, we analyze Range-GAN's performance in a single constraint application. In this paper, we train Range-GAN using the volume ratio and aspect ratio labels separately. For the aspect ratio case, we present some of our results visually in Fig.~\ref{fig:ratio_res}-Left. It can be visually confirmed that, as the input aspect ratio condition increases, the ratio between the fuselage length and wingspan increases, which demonstrates Range-GAN is working as intended. The same visual confirmation is available in the volume ratio dataset. The samples of Range-GAN conditioned on different volume ratio ranges are presented in Fig.~\ref{fig:ratio_res}-Right. In this case as well, one can visually confirm that Range-GAN is performing as expected.
With the results now visually confirmed, we can move on to measure the performance of the models using our metrics. To demonstrate the performance of the model across the label space, we compute the satisfaction metric for input ranges of length 0.05, 0.1, and 0.2 spanned from one side of the label space to the other for 100 input condition ranges spanned uniformly across the label space. For each condition, we generate 2,000 samples and use the predicted labels from the estimator to compute the condition satisfaction. Further, we show the satisfaction calculated for the same ranges in the data. The curves representing data are essentially demonstrating the probability of meeting the input conditions to the generator if we were to randomly sample airplanes from the dataset. The results of this are presented in Fig.~\ref{fig:satisfaction} and Table.~\ref{table:summary}. As evident, Range-GAN is performing very well when it comes to the single constraint range conditioning. Further, we observe that as the input constraint range becomes narrower, the performance of Range-GAN declines. This is expected as, when the range becomes more and more restrictive, it becomes more difficult for the generator to meet the input conditions. Finally, if the input range's bounds converge to a single point, it becomes practically impossible to meet the exact input condition up to machine precision~(\ie, $\epsilon_{machine}$).
We can then move onto measuring the uniformity of the label distribution of output samples. We use the quadratic entropy metric discussed before. To measure this, we take a similar approach as we did for satisfaction. We measure the quadratic entropy at different input conditions across the condition space. This time, however, we train Range-GAN without the uniformity loss to establish a baseline measurement for the effectiveness of the uniformity loss. We measure the quadratic entropy of the labels of 1,000 generated samples that meet the condition at every point in the conditioning space. Because uniformity is only relevant inside the acceptable range, we measure the uniformity only for samples that meet the input condition. For the sake of brevity, we only present the results for the aspect ratio case study. In Fig.~\ref{fig:uniformity}, we observe that the entropy increases significantly after introducing the uniformity loss. This demonstrably indicates the effectiveness of the introduced uniformity loss term. Furthermore, we observe that the difference between the label distributions' entropy becomes more pronounced with an increase in the size of the input condition range. This means that, in more broad input conditions, the GAN with a range loss alone can simply generate samples closer to each other within the acceptable range without any incentive to cover the entire acceptable label space without the uniformity loss. We visually demonstrate this in Fig.~\ref{fig:dist}, which shows one example of how the loss term improves the output distribution to allow for uniform coverage of the label space.

\subsection{Multi-Constraint Case Study}
\begin{figure*}[ht!]
\centering
\includegraphics[width=1.3\columnwidth,height=2.5in]{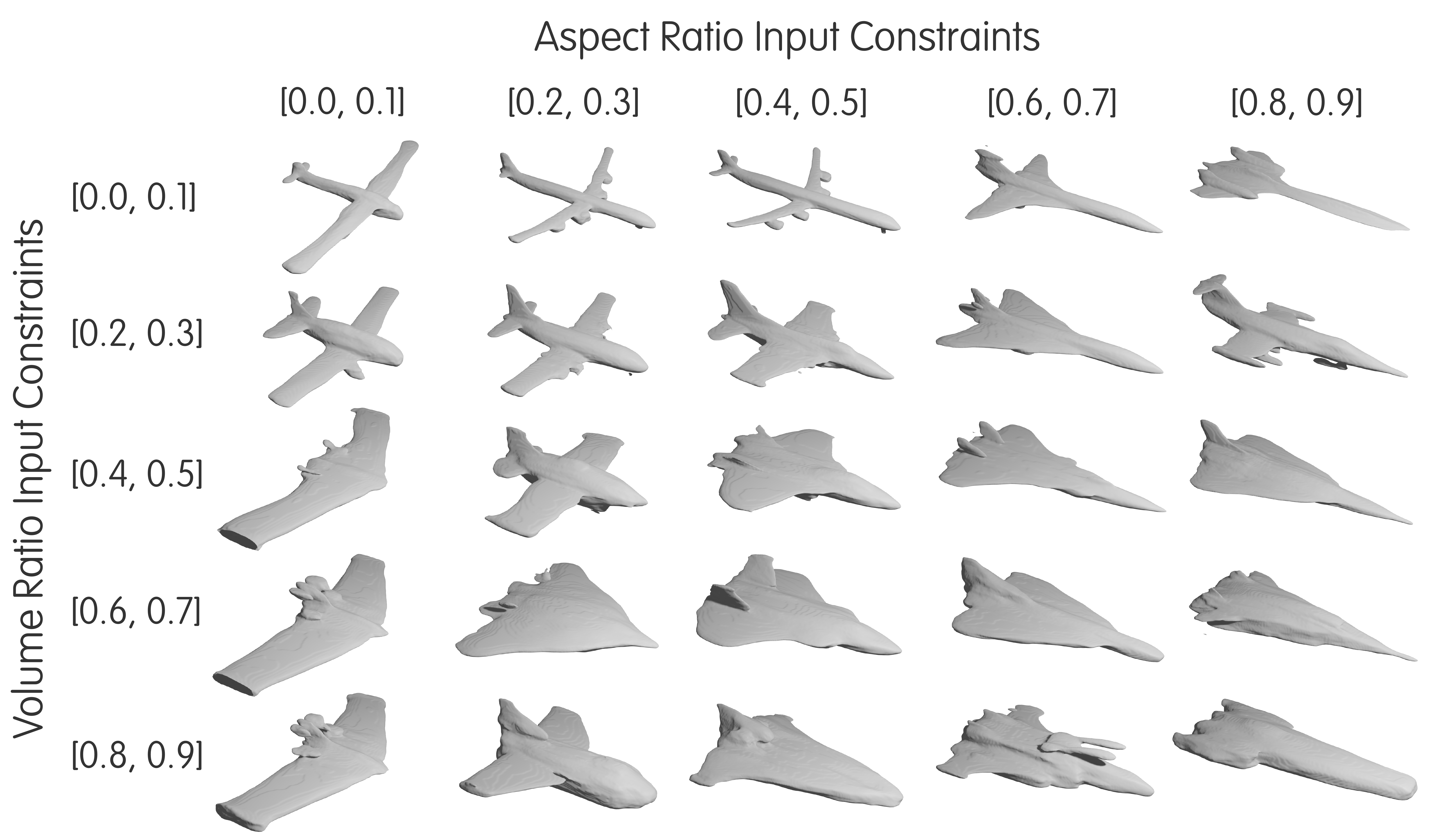}
\vskip -0.1in
\caption{Example of Range-GAN outputs with different volume ratio and aspect ratio conditions. The input range conditions are presented above each column and to the left of each row. The images are to scale so models that appear larger occupy a larger volume, hence a larger volume ratio.}
\label{fig:mo_res}
\end{figure*}

\begin{figure}[ht!]
\centering
\includegraphics[width=0.7\columnwidth]{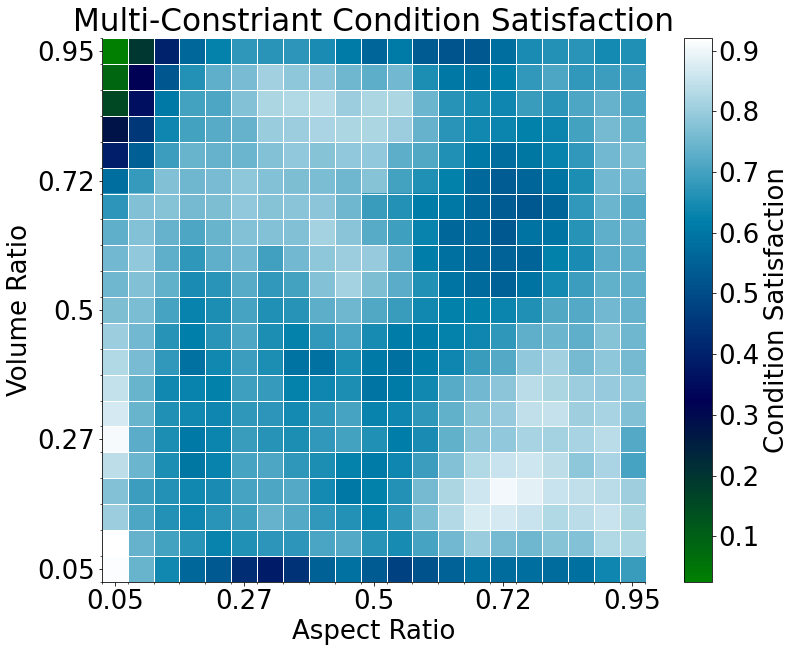}
\vskip -0.1in
\caption{The performance of Range-GAN in input condition satisfaction in both volume and aspect ratio. The Values indicated on the x and y axes of the plots represent the center of the input range conditions with the range size of 0.1. The white grid lines indicate the start and end of the input condition ranges.}
\label{fig:mo_sat}
\end{figure}

Now that we have established the effectiveness of our approach in single constraint range conditioning, we will demonstrate that Range-GAN can be applied in multi-constraint range conditioning as well. We do this by passing range conditions for both the aspect ratio and the volume ratio to the generator. For the case of multi-constraint Range-GAN, we present some of our results visually in Fig.~\ref{fig:mo_res}. This figure demonstrates samples generated by Range-GAN that meet the input conditions since only one sample is presented for each input condition. Regardless, the visual trend seen here is present in generated samples overall. As evident in this image, the two trends seen before for the single constraint cases are present together, visually demonstrating the effectiveness of the Range-GAN in multi-constraint conditioning.
Having established our results visually, we will present the condition satisfaction for the multi-constraint case study. Given the difficulty of presenting the higher dimensional data, we will only present our results for the range size of 0.1 for both constraints. Similar to the single constraint case before, we compute the satisfaction for 2,000 generated samples at each condition and calculate the satisfaction for 20 conditions spanned uniformly across each label space. The results of this analysis are presented in Fig.~\ref{fig:mo_sat} and Table.~\ref{table:summary}. Comparably to the single constraint case, Range-GAN is capable of conditioning effectively despite the very limited distribution of the data labels~(Fig.~\ref{fig:overall_dist}), which demonstrates the fact that the approaches discussed in Range-GAN can be expanded to multi-constraint applications effectively. We do, however, observe that the overall satisfaction is lower in the multi-constraint case than in the single constraint cases. This is expected, as the complexity of the task quadruples in going from a single constraint to two constraints. Another important observation is that Range-GAN has failed to produce low aspect ratio and high volume ratio samples, as can be seen in the top left corner of the Range-GAN satisfaction plot. This is not necessarily unexpected, as the data is practically non-existent in that location. A low aspect ratio requires that the generated aircraft have a significantly smaller fuselage compared to its wingspan, and, given that the fuselage typically contains most of the aircraft's volume, it becomes very difficult, in fact practically impossible, to generate large numbers of airplanes with such properties.
\begin{table}[h!]
\vskip -0.1in
\caption{Experimental results on condition satisfaction (mean and standard deviation over the entire condition space) with estimator prediction}
\label{table:summary}
\begin{tabular}{ccc}
    \toprule
    Condition Variable & Range Size & Condition Satisfaction\\
    \midrule
    Volume Ratio& 0.1&0.9131$\pm$0.0309\\
    Volume Ratio& 0.2&0.9841$\pm$0.0102\\
    Aspect Ratio& 0.1&0.8723$\pm$0.03806\\
    Aspect Ratio& 0.2&0.9687$\pm$0.0112\\
    Both&0.1&0.6921$\pm$0.1102\\
    Both&0.2&0.9294$\pm$0.0432\\
    \bottomrule
\end{tabular}
\vskip -0.2in
\end{table}

\subsection{Effects of Data Augmentation}
\begin{figure}[ht!]
\vskip -0.2in
\centering
\includegraphics[width=1\columnwidth]{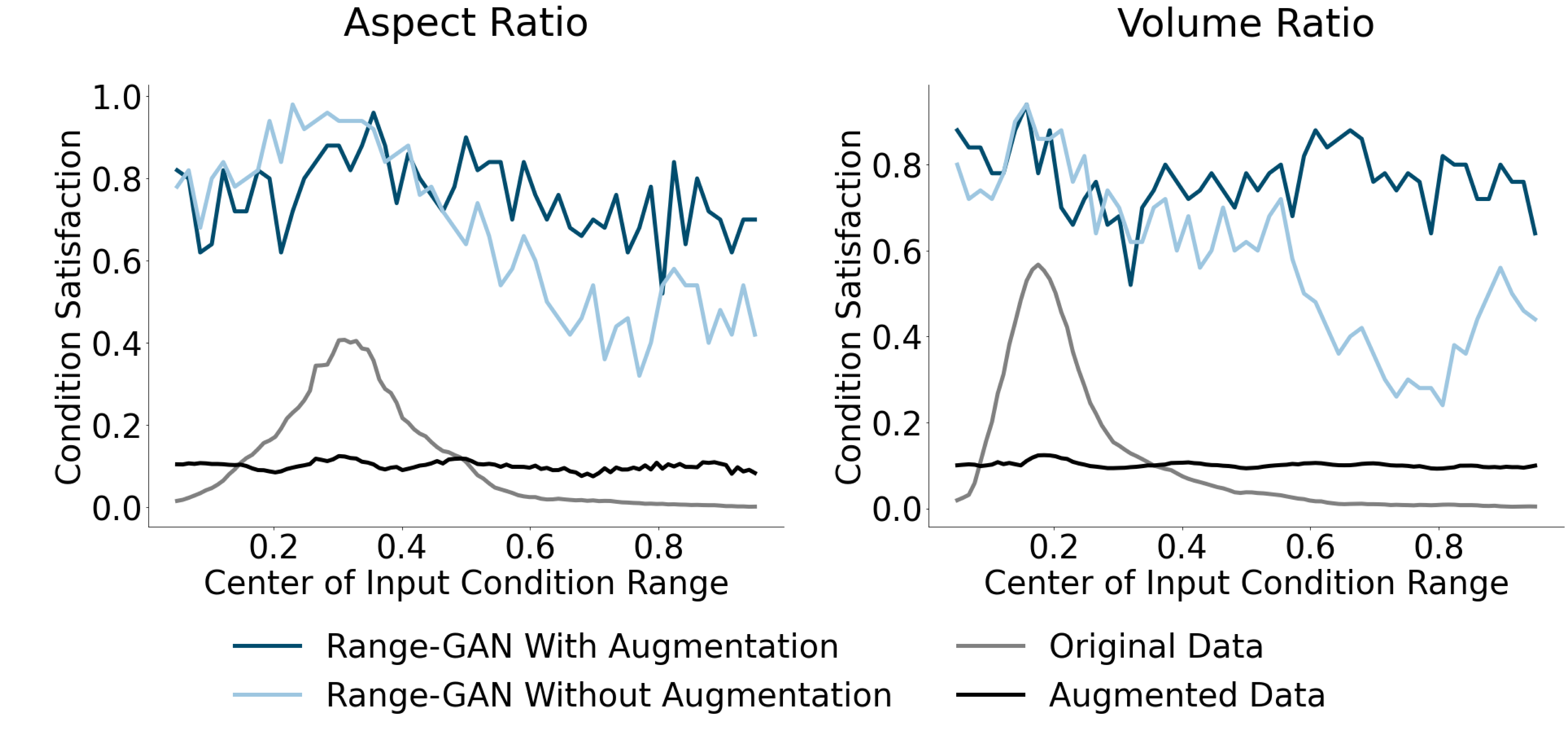}
\vskip -0.1in
\caption{The real-world performance of Range-GAN in input condition satisfaction in both volume and aspect ratio compared to the same metric in the data distribution both before and after data augmentation for a range size of 0.1. The x-axis on the plots represents the center of the input range condition.}
\vskip -0.2in
\label{fig:real_sat}
\end{figure}

So far, we have evaluated the performance of our models based on the DNN estimator predictions of performance. In this section, we will discuss the real measured performance of the generated samples by generating samples and reconstructing them using the implicit decoder, and measuring their volume and aspect ratios. This task is computationally expensive, which is why it is not practical to show these results for all of our work in this manner. Further, the methodology surrounding Range-GAN assumes that an accurate estimator is being used, which is crucial for the success of Range-GAN; measuring the performance based on the estimator shows how well the range loss works in guiding Range-GAN towards the correct labels. Nevertheless, the real-world implications of using DNN-based approximate estimators must be discussed. Additionally, we do this to show that the proposed data-augmentation improves the estimator, hence the real-world performance of Range-GAN. First, we present the real-world performance of the models with and without augmentation for the single constraint case-studies at 50 conditions uniformly spanned across the condition space for a range size of 0.1. For every condition, we generate and construct 50 3D samples and measure their real-world performance. The actual satisfaction are presented for both volume and aspect ratios in Fig.~\ref{fig:real_sat} and Table.~\ref{table:aug}. As is evident, the real-world performance of Range-GAN is worse than the performance of Range-GAN based on the estimator prediction. Further, we can see that the data after augmentation is relatively uniformly distributed across the label space, which has helped improve the DNN estimator significantly. We find that Range-GAN's real performance is more consistent in both volume and aspect ratio conditioning when augmentation is applied, while the Range-GAN without augmentation is more inconsistent in performance with significant dips in satisfaction. This proves the original claim that the DNN estimator would struggle in sparse regions, which we see specifically in the higher end of both volume and aspect ratios; the un-augmented Range-GAN performs poorly here, while the augmented Range-GAN does better. It is important to mention that the computational cost of augmentation is significant, as shapes have to be reconstructed to calculate their labels. The benefits, however, seem significant enough to justify such costs.
\begin{table}[h!]
\vskip -0.3in
\caption{Experimental results on condition satisfaction with a range size of 0.1 (mean and standard deviation over the entire condition space) with exact performance calculation}
\label{table:aug}
\label{table:summary}
\begin{tabular}{cccc}
    \toprule
    Condition Variable & Condition Satisfaction & Augmented\\
    \midrule
    Volume Ratio&0.5823$\pm$0.1861&No\\
    Volume Ratio&0.7682$\pm$0.0763&Yes\\
    Aspect Ratio&0.6745$\pm$0.1926&No\\
    Aspect Ratio&0.7580$\pm$0.0896&Yes\\
    \bottomrule
\end{tabular}
\vskip -0.2in
\end{table}

\subsection{Limitations and Future work}

The approach taken in Range-GAN has some notable limitations. First, is the fact that the method is heavily reliant on a differentiable estimator, the quality of which will determine the performance of Range-GAN. Therefore, great care must be taken when selecting the estimator, and exact estimators are always preferred. Unfortunately, in most design applications exact estimators are not readily differentiable, and even if they were, the evaluation speed is often quite slow and thus impractical. This means that, more often than not, we are bound to use a DNN-based estimator. Therefore, the most significant limitation in this approach to conditioning in continuous spaces is the estimator. Consequently, approaches in improving estimator models for more accurate guidance of the GAN are very important in this application. In the future, we intend to develop improved methods for obtaining highly accurate data-driven estimators that can mimic high-fidelity physics simulations and other exact estimators to create more consistency in results between Range-GAN's estimator predicted performance and its real-world performance.


Besides, it is important to note that the methodology presented in this paper is extendable to other domains of design, where data is available and labeled. In the future, Range-GAN can be applied to any domain to allow for design synthesis under continuous constraints, which, given the current lack of such tools, is an important contribution to the field of data-driven design synthesis.

\section{Conclusion}
In this paper, we introduced an approach that allows for data-driven design synthesis under range constraints in continuous label spaces. Range-GAN is the first to address this problem. To achieve this, we introduced a novel architecture that uses a pre-trained estimator to guide Range-GAN towards achieving proper conditioning through a novel loss function, the `range loss'. We demonstrated the effectiveness of this approach in both single constraint design and multi-constraint design using a 3D shape synthesis example to generate airplane models. We showed that Range-GAN can successfully generate samples that meet the input conditions, even when the dataset is extremely sparse in certain parts of the label space, achieving more than 80\% satisfaction of input range condition.

Another aspect of conditioning under range constraint is the output distribution of the generated samples' labels. In this paper, we developed an approach that encourages uniform coverage of the label space in the acceptable condition range. 
To achieve this, we introduced a novel loss function, the `uniformity loss', to encourage uniform coverage of the input constraint range. We demonstrated this loss function's effectiveness at encouraging uniform coverage of the label space by comparing Range-GAN results with and without this loss, finding that the loss function is highly effective and more than doubles the label entropy.

We also analyzed how Range-GAN can be improved by label-aware self-augmentation of the data. We showed this by augmenting the data using the Range-GAN's own generated samples to add more samples to sparse regions of the label space, enabling us to re-train the DNN-based estimator and Range-GAN using this augmented data to improve the performance of Range-GAN significantly. We show that the label-aware self-augmentation leads to an average improvement of 14\% on the constraint satisfaction for generated 3D shapes.

This work laid the foundation for data-driven inverse design problems where we consider range constraints and there are sparse regions in the condition space. Both situations are common in engineering design scenarios. While we validated our proposed model on a 3D shape synthesis example, the method is not restricted to this application. For example, by replacing the latent vector with parameters of unit cell shapes, this model can also help address the inverse design problem of cellular structures~\cite{wang2021ih}.

\bibliographystyle{asmems4}


%
% ...

\bibliography{asme2e}


\end{document}